\documentclass[letterpaper, preprint, paper,10pt]{AAS}	

\usepackage{bm}
\usepackage{amsmath}
\usepackage{subfigure}
\usepackage[colorlinks=true, pdfstartview=FitV, linkcolor=black, citecolor= black, urlcolor= black]{hyperref}
\usepackage{overcite}
\usepackage{footnpag}			      	

\usepackage{tikz} 
\usetikzlibrary{positioning, arrows.meta}
\usepackage{amsmath}
\usepackage{amssymb}
\PaperNumber{26-826}

\makeatother

\begin{document}

\title{Early Prediction of Satellite Collision Probability
Using a Hybrid TCN–Transformer Model for a
CDM-Based Conjunction Analysis Framework}

\author{Rabia T. Tok\thanks{Graduate Researcher, Space Mission Analysis and Design Group, TÜBİTAK UZAY, tuylek17@itu.edu.tr},  
Burak Yağlıoğlu\thanks{Senior Chief Researcher, Space Mission Analysis and Design Group, TÜBİTAK UZAY, burak.yaglioglu@tubitak.gov.tr},
Enes Dağ\thanks{Chief Researcher, Space Mission Analysis and Design Group, TÜBİTAK UZAY, enes.dag@tubitak.gov.tr},
\ and Emre O. Kahya\thanks{Professor, Department of Physics Engineering, Istanbul Technical University, eokahya@itu.edu.tr}
}

\maketitle{}

\begin{abstract}
The rapid expansion of operational satellites and orbital debris has increased the frequency of close approach events in low Earth orbit (LEO), creating a higher operational burden for satellite operators. This problem is especially critical for satellites using electric propulsion, where low-thrust maneuver capability imposes additional time constraints on collision avoidance planning. In current practice, Conjunction Data Messages (CDMs), defined by the Consultative Committee for Space Data Systems (CCSDS), provide relative state, covariance, miss distance, time of closest approach, and probability of collision (PoC) information for conjunction assessment. However, the nonlinear propagation of orbital uncertainties and the sensitivity of PoC to covariance evolution make the interpretation of sequential CDMs challenging. This study proposes a learning-based framework for early prediction of satellite conjunction risk by estimating the PoC expected in the subsequent CDM update of the same close approach event. In the proposed methodology, an Unscented Transform-based propagation and backpropagation framework is first used to evaluate the sensitivity of the collision risk metric to CDM parameters. In addition, Principal Component Analysis is applied to the numerical CDM parameters to identify the features most relevant to PoC variation. The results obtained from the sensitivity analysis and PCA are then used to justify the selected raw CDM parameters and to construct derived metrics representing relative motion, encounter geometry, and covariance-related uncertainty. Using the resulting sequential enriched conjunction dataset, a hybrid Temporal Convolutional Network (TCN)--Transformer model is trained to learn the temporal evolution of conjunction risk. The framework is applied to CDMs received and analyzed within T{\"U}B{\.I}TAK UZAY, demonstrating its potential for earlier and more consistent operational risk evaluation for LEO satellite conjunctions.

\end{abstract}

\section{Introduction}
The increasing population of operational satellites and orbital debris has significantly intensified congestion in Earth orbit, particularly in low Earth orbit (LEO), where large constellations are actively deployed. This growth increases the frequency of close approach events and creates a higher operational burden for satellite operators.\cite{ref1} Even small debris objects may cause severe damage due to high relative orbital velocities; therefore, timely and reliable conjunction risk assessment has become an essential part of satellite operations.\cite{ref2,ref3,ref4}

In current operational practice, Conjunction Data Messages (CDMs) provide the main information used to assess close approach events, including relative state vectors, covariance matrices of the primary and secondary objects, miss distance, time of closest approach (TCA), and estimated probability of collision (PoC).\cite{ref5} A conjunction event may be represented by multiple CDMs generated at different times before TCA. Although later CDMs generally provide more updated and reliable information, waiting for the final CDM reduces the available time for analysis, maneuver planning, command preparation, and maneuver execution. This time constraint is especially critical for satellites using electric propulsion systems. Since electric propulsion provides low thrust, collision avoidance maneuvers require longer execution times compared with chemical propulsion. Therefore, an early understanding of how the collision risk may evolve is important for initiating operational planning before the final CDM update becomes available.\cite{ref6,ref7} In this context, predicting the PoC of a future CDM can provide additional support for early conjunction risk assessment. The interpretation of consecutive CDMs is challenging because the reported PoC is strongly affected by the time evolution of orbital uncertainty. When the satellite state is numerically propagated, the associated covariance matrix generally expands due to nonlinear orbital dynamics, dynamical perturbations, and initial state uncertainty. This covariance growth changes the uncertainty ellipsoid around the nominal satellite position and directly affects the collision probability calculation at the encounter epoch. For this reason, there is a need for predictive frameworks that can learn the temporal evolution of uncertainty and collision risk from sequential CDM before the next CDM update becomes available.

This study proposes a physically grounded and data-based framework for early prediction of LEO satellite conjunction risk. The contribution of CDM parameters to the collision risk metric is quantified using the Unscented Transform. Through nonlinear state and covariance propagation, the relative influence of position, velocity, and uncertainty components on PoC evolution is analyzed. Based on the resulting sensitivity analysis and PCA , a structured dataset generation methodology is developed by calculating derived physical metrics and by using raw data.  A hybrid TCN--Transformer neural network is trained to estimate the PoC value expected in the subsequent CDM update with this dataset. The proposed framework supports earlier identification of potentially hazardous conjunctions and provides operational support for maneuver planning before the next CDM becomes available. The methodology is applied to CDMs received and analyzed within T\"UB\.ITAK UZAY, demonstrating its practical relevance for satellite missions.

\section{Conjunction Risk Assessment Framework}
\subsection{CDM Based Conjunction Assessment}
\begin{figure}[htbp]
\centering
	\centering\includegraphics[width=4.5in]{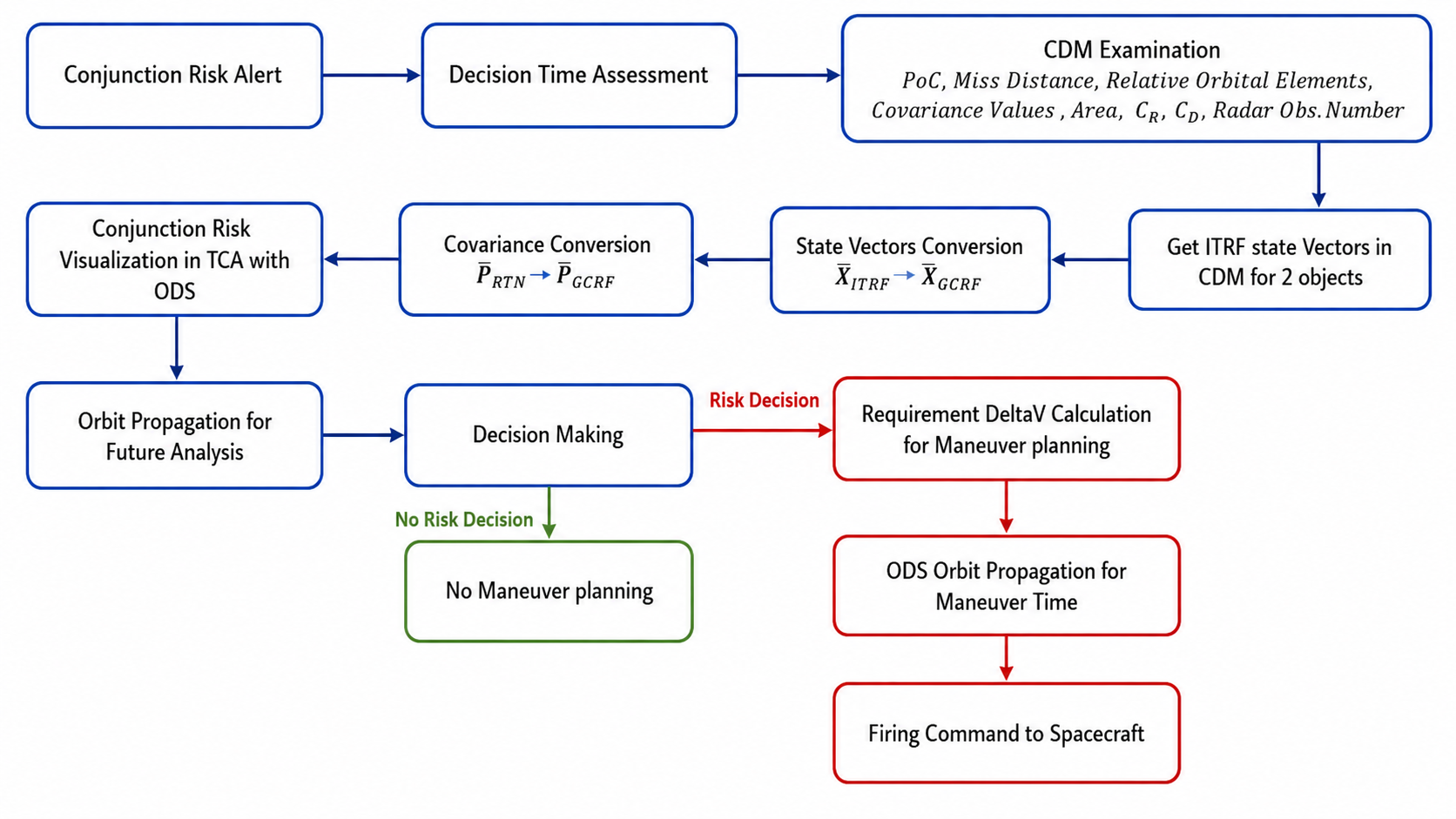}
    \caption{Collision Avoidance Operation Framework.}
    \label{fig:operationalFlow}
\end{figure}

Conjunction risk assessment is evaluated when a CDM is received for a TCA between two space objects which may include active satellites, rocket bodies, debris fragmentation. A CDM provides an updated description of a close approach event, including the creation time of the message, the TCA, miss distance, state vectors of primary and secondary objects at ITRF, relative position and velocity components at RTN, calculated PoC, covariance matrices of both objects at RTN, physical object properties, observation related information, and the modeling assumptions used during conjunction assessment. The operational flow is shown in Figure~\ref{fig:operationalFlow}.
After a conjunction risk alert is received, all parameters in incoming CDM are evaluated through the available decision time until TCA. The state vectors and the covariance matrices are converted into the GCRF frame to evaluate the encounter geometry and uncertainty distribution. These parameters are propagated by using orbital dynamics system to enable the reliability of the maneuver decision making process.\cite{ref8} Although the maneuver decision is not based solely on PoC, it is used as a supporting risk indicator together with the encounter geometry, covariance information, miss distance, and operational constraints. In this study, conjunctions with $P_c \geq 10^{-6}$ are generally treated as risky, while those with $P_c < 10^{-6}$ are considered non-risky. Non-risky events do not require maneuver planning, whereas risky events require collision avoidance maneuver planning with the computation of the required $\Delta V$.

Event which is related to close approach can be more than single CDM; however, this process is mainly based on the latest available CDM and does not directly estimate how the collision risk may evolve in the next CDM update. Therefore, a predictive framework is required to support earlier decision making before the final CDM is available. 

\subsection{PoC Representation}

In each CDM, the reported PoC corresponds to the collision probability evaluated using the state, covariance, and encounter geometry information contained in that message. Since these quantities may be updated from one CDM to another, the PoC value reflects the instantaneous risk estimate associated with the available conjunction information. In encounter geometry, the three-dimensional relative covariance is projected onto the two-dimensional encounter plane, which is perpendicular to the relative velocity vector. The projected covariance is represented by an uncertainty ellipse, whereas the hard-body collision region is represented by a circular area defined by the combined object radius. The probability of collision is calculated by integrating the two-dimensional Gaussian probability density function over the hard-body region in the encounter plane, as expressed in Eq.~(\ref{eq:POC}).
\begin{equation}
\label{eq:POC}
P_c = \frac{1}{2\pi \sigma_x \sigma_y}
\int_{-OBJ}^{OBJ}
\int_{-\sqrt{OBJ^2 - x^2}}^{\sqrt{OBJ^2 - x^2}}
\exp\left(
-\frac{1}{2}
\left[
\left(\frac{x - x_m}{\sigma_x}\right)^2 +
\left(\frac{y - y_m}{\sigma_y}\right)^2
\right]
\right)
dy \, dx
\end{equation}

$\sigma_x$ and $\sigma_y$ represent the standard deviations along the principal axes of the encounter plane. $x_m$ and $y_m$ denote the components of the projected miss distance. $OBJ$ is the combined object radius. $x$ and $y$ are the coordinates in the encounter plane.\cite{ref9} The covariance information provided in each CDM represents the uncertainty of the object state at the encounter epoch. In three-dimensional space, this uncertainty is interpreted as a covariance cloud or uncertainty ellipsoid around the nominal position of the object. PoC represents the collision probability of two space objects based on this uncertainty distribution. Equation~(\ref{eq:POC}) shows that the PoC is computed using the information available in the current CDM. However, if a subsequent CDM is obtained for the same close approach event, the PoC value of the next CDM update is predicted using the neural network model to support operational process.

\section{Methodology}
The dataset methodology is constructed by calculating how strongly the parameters in the CDMs are related to the PoC and to what extent these parameters affect the covariance. In addition, since the probability of collision is computed using uncertainty information, a dataset structure is developed by analyzing how the covariance ellipsoid of the corresponding object changes during orbit propagation. Sensitivity analysis is used to analyze how the covariance ellipsoid changes as the satellite propagates, and Principal Component Analysis is used to evaluate the relationship between each parameter included in the CDM and the PoC.

\subsection{Orbit Propagation with UT}
\begin{figure}[htbp]
\centering
	\centering\includegraphics[width=4.5in]{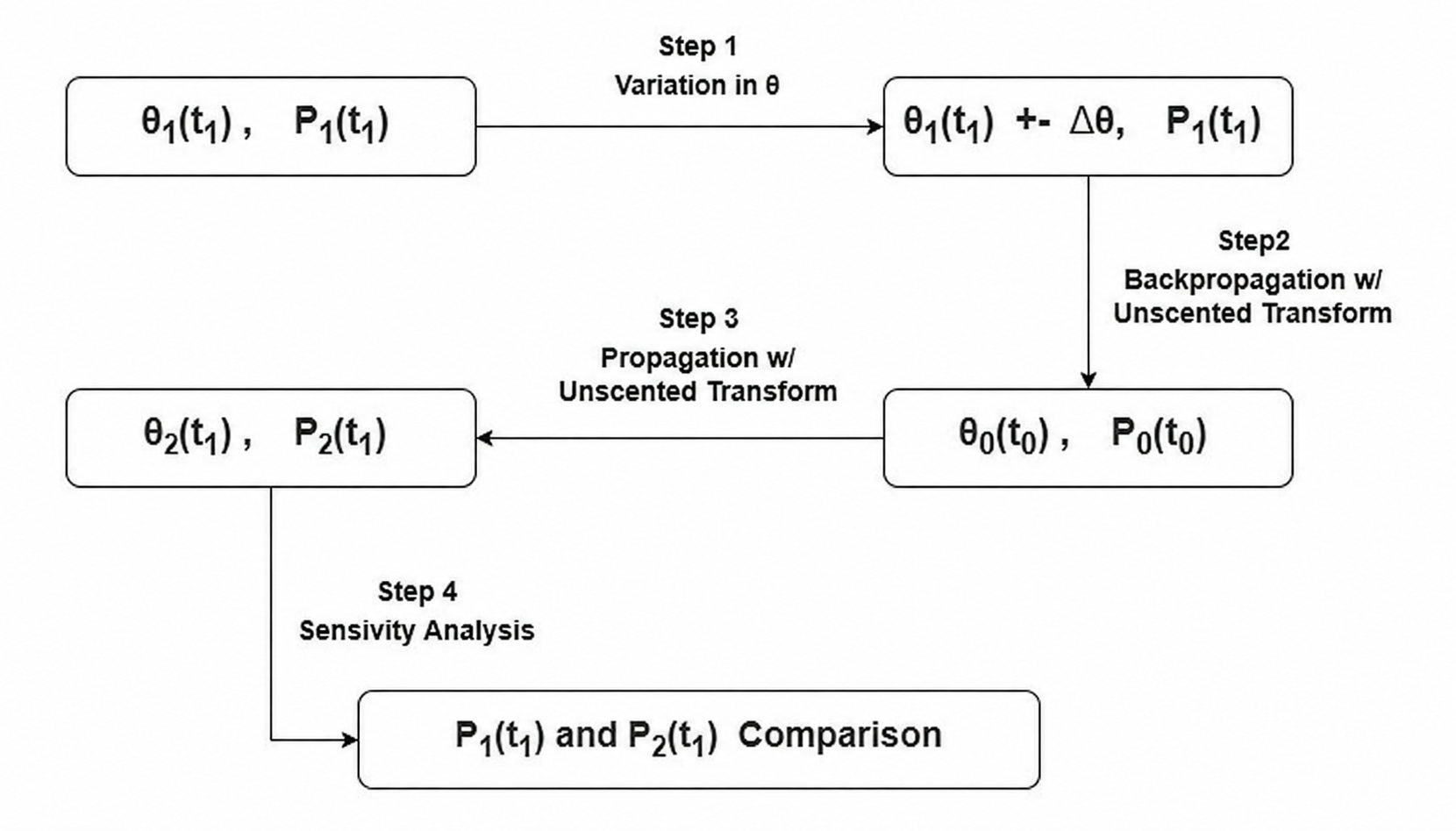}
    \caption{Dataset Generation Workflow.}
    \label{fig:datasetWorkflow}
\end{figure}

Instead of using random perturbations, the proposed framework applies controlled variations to selected physical and dynamical parameters in order to assess their effects on the propagated state and covariance. Within this framework, the Unscented Transform is used to propagate both the state vector and its corresponding covariance through nonlinear orbital dynamics with high accuracy. This approach makes it possible to capture the nonlinear interactions between uncertainty and orbital evolution. Following the propagation step, the resulting covariance matrices are examined using a sensitivity analysis approach to quantify the contribution of each perturbed parameter to the overall uncertainty growth. Figure~\ref{fig:datasetWorkflow} illustrates the steps of the dataset construction methodology based on sensitivity analysis.

The parameter vector $\boldsymbol{\theta}$ represents the main inputs required for orbit propagation, including the satellite state and the physical parameters that affect the propagation dynamics. These parameters are defined in Eq.~\eqref{eq:theta_vector}.
\begin{equation}
\boldsymbol{\theta} =
\begin{bmatrix}
x & y & z & v_x & v_y & v_z & C_d & C_r & A
\end{bmatrix}^T
\label{eq:theta_vector}
\end{equation}
where $(x, y, z)$ and $(v_x, v_y, v_z)$ denote the position and velocity components of the satellite, respectively. The additional parameters $C_d$, $C_r$, and $A$ represent the drag coefficient, solar radiation pressure coefficient, and effective cross-sectional area.

As shown in Figure~\ref{fig:datasetWorkflow}, the initial parameter vector $\boldsymbol{\theta}_1$ and covariance matrix $\mathbf{P}_1$ at time $t_1$ are obtained directly from real CDM. The main objective of this procedure is to introduce a controlled variation into the system and to evaluate how this variation affects the covariance when the state is propagated backward and then forward to the same epoch $t_1$. This allows the effect of each selected parameter on the propagated covariance to be analyzed independently. The UT is utilized during dataset generation to perform state and covariance propagation and backpropagation, enabling the identification of CDM parameters that significantly influence the PoC. State and covariance propagation is performed using an RK8 integrator. As shown in Equation~\eqref{eq:full_dynamics}, the dynamical model used in the propagation process includes acceleration components acting on the satellite: central two-body gravitational acceleration, Earth gravity model perturbations, atmospheric drag acceleration, solar radiation pressure acceleration, and third-body gravitational perturbations.\cite{ref10, ref11}
\begin{equation}
\ddot{\mathbf{r}}(t) =
\mathbf{a}(t) =
\mathbf{a}_{2body}
+ \mathbf{a}_{EGM}
+ \mathbf{a}_{Drag}
+ \mathbf{a}_{SRP}
+ \mathbf{a}_{3rdBody}
\label{eq:full_dynamics}
\end{equation}
In the first step, a variation $\Delta\boldsymbol{\theta}$ is applied to only one selected parameter in the $\boldsymbol{\theta}$ vector, while all remaining parameters are kept unchanged. The perturbed state and covariance are then backpropagated from $t_1$ to $t_0$ using the Unscented Transform. After this step, the obtained state and covariance at $t_0$ are propagated forward again to $t_1$. The resulting covariance matrix $\mathbf{P}_2(t_1)$ is then compared with the original covariance matrix $\mathbf{P}_1(t_1)$. The change between these covariance matrices is used to quantify the sensitivity of the covariance to the selected parameter.

\subsubsection{Step 1}
Within the CDM parameter set, a single parameter is selected and perturbed by applying a $1\%$ controlled variation, while all remaining parameters are held constant. This procedure is repeated separately for each parameter in order to analyze the individual contribution of each variable to the propagated covariance. If the x component of the position vector in the CDM is increased by $1\%$, the new state vector is expressed in Equations~\eqref{eq:xIncrease} and ~\eqref{eq:state}.
\begin{equation}
x_1^{*} = x_1 + \Delta x, 
\quad \Delta x = 0.01 x_1
\label{eq:xIncrease}
\end{equation}

\begin{equation}
\mathbf{x}_1^{*}(t_1)
=
\begin{bmatrix}
\mathbf{r}_1^{T} & \mathbf{v}_1^{T}
\end{bmatrix}^{T}
=
\begin{bmatrix}
x_1^{*} & y_1 & z_1 & v_{x1} & v_{y1} & v_{z1}
\end{bmatrix}^{T}
\label{eq:state}
\end{equation}
The uncertainty associated with the modified state vector is represented by the covariance matrix  $\mathbf{P}_1(t_1)$. This covariance matrix is directly obtained from the corresponding CDM and is used  as the initial covariance matrix in the Unscented Transform. During Step 1, only the selected state parameter is perturbed, while the covariance matrix $\mathbf{P}_1(t_1)$ is kept unchanged. In the CDM, the covariance components are provided in the RTN frame. Therefore, before the backpropagation and propagation steps, all covariance matrices are transformed into a common inertial reference frame, namely GCRF, in order to ensure consistency with the state propagation model. The CDM covariance matrix in the RTN frame is expressed as
\begin{equation}
\mathbf{P}^{RTN}_1(t_1) =
\begin{bmatrix}
C_{RR} & C_{RT} & C_{RN} & C_{R\dot{R}} & C_{R\dot{T}} & C_{R\dot{N}} \\
C_{RT} & C_{TT} & C_{TN} & C_{T\dot{R}} & C_{T\dot{T}} & C_{T\dot{N}} \\
C_{RN} & C_{TN} & C_{NN} & C_{N\dot{R}} & C_{N\dot{T}} & C_{N\dot{N}} \\
C_{R\dot{R}} & C_{T\dot{R}} & C_{N\dot{R}} & C_{\dot{R}\dot{R}} & C_{\dot{R}\dot{T}} & C_{\dot{R}\dot{N}} \\
C_{R\dot{T}} & C_{T\dot{T}} & C_{N\dot{T}} & C_{\dot{R}\dot{T}} & C_{\dot{T}\dot{T}} & C_{\dot{T}\dot{N}} \\
C_{R\dot{N}} & C_{T\dot{N}} & C_{N\dot{N}} & C_{\dot{R}\dot{N}} & C_{\dot{T}\dot{N}} & C_{\dot{N}\dot{N}}
\end{bmatrix}
\label{eq:cdm_covariance_rtn}
\end{equation}

\subsubsection{Step 2}
The modified state is then backpropagated in time for one day using a fixed integration step size of 240 seconds. The Unscented Transform generates sigma points around the modified state and covariance, and each sigma point is propagated backward through the nonlinear orbital dynamics using the RK8 integrator. Therefore, the UT does not directly propagate only the mean state; instead, it propagates the sigma-point distribution through the RK8-based nonlinear dynamics and reconstructs the predicted mean state and covariance at the previous epoch. The nonlinear state model used for the propagation is written as
\begin{equation}
\dot{\mathbf{x}}(t) =
\begin{bmatrix}
\dot{\mathbf{r}}(t) \\
\dot{\mathbf{v}}(t)
\end{bmatrix}
=
\begin{bmatrix}
\mathbf{v}(t) \\
\mathbf{a}(t)
\end{bmatrix}
\end{equation}

\begin{equation}
\dot{\mathbf{x}}(t) = \mathbf{f}(t,\mathbf{x}(t))
\end{equation}
$\mathbf{x}(t)$ denotes the satellite state vector. For the numerical integration of this nonlinear system, the RK8 scheme advances the solution from $t_k$ to $t_{k+1}$ as
\begin{equation}
\mathbf{x}_{k+1}
=
\mathbf{x}_{k}
+
h
\sum_{i=1}^{s}
b_i \mathbf{k}_i 
\end{equation}
$h$ is the integration step size, $s$ is the number of RK8 stages, and $b_i$ denotes the RK8 weights. The intermediate stages are computed as
\begin{equation}
\mathbf{k}_1
=
\mathbf{f}(t_k,\mathbf{x}_k)
\end{equation}
\begin{equation}
\mathbf{k}_i
=
\mathbf{f}
\left(
t_k + c_i h,
\mathbf{x}_k
+
h
\sum_{j=1}^{i-1}
a_{ij}\mathbf{k}_j,
\right),
\qquad i=2,\ldots,s 
\end{equation}

where $a_{ij}$, $b_i$, and $c_i$ are the RK8 coefficients.\cite{ref12} In the backpropagation step, the same RK8 formulation is used with a negative time direction. The previous epoch is defined as
\begin{equation}
t_0 = t_1 - 86400 \ \mathrm{s}
\end{equation}
Within the UT-based backpropagation, the modified state vector and covariance at $t_1$ are denoted as $\mathbf{x}_1^{*}(t_1)$ and $\mathbf{P}_1(t_1)$. $\mathbf{x}_1^{*}(t_1)$ is used as the initial mean state vector of the Unscented Transform. The mean vector at the initial epoch is defined as
\begin{equation}
\bar{\mathbf{x}}(t_1) = \mathbf{x}_1^{*}(t_1)
\label{eq:modified_mean_state}
\end{equation}
A set of $2n+1$ sigma points is then generated around this modified mean state in order to represent the uncertainty distribution of the state:
\begin{equation}
\boldsymbol{\chi}_0(t_1) =
\bar{\mathbf{x}}(t_1)
=
\mathbf{x}_1^{*}(t_1)
\label{eq:sigma0_backprop}
\end{equation}
\begin{equation}
\boldsymbol{\chi}_i(t_1)
=
\bar{\mathbf{x}}(t_1)
+
\left[
\sqrt{(n+\lambda)\mathbf{P}_1(t_1)}
\right]_i,
\qquad i=1,\ldots,n
\label{eq:sigma_plus_backprop}
\end{equation}
\begin{equation}
\boldsymbol{\chi}_{i+n}(t_1)
=
\bar{\mathbf{x}}(t_1)
-
\left[
\sqrt{(n+\lambda)\mathbf{P}_1(t_1)}
\right]_i,
\qquad i=1,\ldots,n
\label{eq:sigma_minus_backprop}
\end{equation}
where $n$ is the state dimension. Since the satellite state vector consists of three position and three velocity components, the state dimension is taken as 6. The scaling parameter $\lambda$ is defined as

\begin{equation}
\lambda = \alpha^2(n+\kappa)-n 
\end{equation}

The corresponding mean and covariance weights are given by
\begin{equation}
W_0^{(m)}
=
\frac{\lambda}{n+\lambda}
\end{equation}
\begin{equation}
W_0^{(c)}
=
\frac{\lambda}{n+\lambda}
+
(1-\alpha^2+\beta)
\end{equation}
\begin{equation}
W_i^{(m)}
=
W_i^{(c)}
=
\frac{1}{2(n+\lambda)},
\qquad i=1,\ldots,2n .
\end{equation}
Each sigma point is then backpropagated from $t_1$ to $t_0$ through the RK8 based nonlinear dynamics
\begin{equation}
\boldsymbol{\chi}_i(t_0)
=
\Phi^{RK8}_{t_1 \rightarrow t_0}
\left(
\boldsymbol{\chi}_i(t_1)
\right),
\qquad i=0,\ldots,2n ,
\end{equation}

where $\Phi^{RK8}_{t_1 \rightarrow t_0}(\cdot)$ denotes the RK8 propagation operator applied backward in time. After all sigma points are propagated to $t_0$, the predicted mean state at the previous epoch is reconstructed as

\begin{equation}
\mathbf{x}_{pred}(t_0)
=
\sum_{i=0}^{2n}
W_i^{(m)}
\boldsymbol{\chi}_i(t_0)
\end{equation}

The predicted covariance matrix at $t_0$ is then obtained from the weighted dispersion of the propagated sigma points around the predicted mean
\begin{equation}
\mathbf{P}_{0}(t_0)
=
\sum_{i=0}^{2n}
W_i^{(c)}
\left(
\boldsymbol{\chi}_i(t_0)-\mathbf{x}_{pred}(t_0)
\right)
\left(
\boldsymbol{\chi}_i(t_0)-\mathbf{x}_{pred}(t_0)
\right)^T 
\end{equation}

The RK8 integrator provides the nonlinear orbital propagation of each sigma point, while the Unscented Transform reconstructs the predicted state and covariance after backpropagation.\cite{ref13}

\subsubsection{Step 3}
After the UT-based backpropagation step, the obtained state and covariance at the previous epoch $t_0$ are propagated forward again to the original CDM epoch $t_1$. In this step, the RK8 integrator is again used as the nonlinear orbital dynamics function inside the Unscented Transform. Therefore, the sigma points generated around the state and covariance at $t_0$ are propagated forward through 
the RK8 based orbital dynamics, and the propagated mean state and covariance are reconstructed at $t_1$. The forward propagation interval is defined as
\begin{equation}
t_1 = t_0 + 86400 \ \mathrm{s}
\end{equation}
The state and covariance obtained at the end of the backpropagation step are used as the initial conditions for forward propagation with same step size.

\begin{equation}
\bar{\mathbf{x}}(t_0) = \mathbf{x}_{pred}(t_0),
\qquad
\mathbf{P}(t_0) = \mathbf{P}_0(t_0)
\end{equation}

The sigma points generated around $\bar{\mathbf{x}}(t_0)$ and $\mathbf{P}_0(t_0)$ are propagated from $t_0$ to $t_1$ using the RK8 propagation operator.

\begin{equation}
\boldsymbol{\chi}_i(t_1)
=
\Phi^{RK8}_{t_0 \rightarrow t_1}
\left(
\boldsymbol{\chi}_i(t_0)
\right),
\qquad i=0,\ldots,2n .
\end{equation}

After all sigma points are propagated to $t_1$, the propagated mean state is reconstructed and the corresponding propagated covariance matrix is obtained.

\begin{equation}
\mathbf{x}_2(t_1)
=
\sum_{i=0}^{2n}
W_i^{(m)}
\boldsymbol{\chi}_i(t_1)
\end{equation}
\begin{equation}
\mathbf{P}_2(t_1)
=
\sum_{i=0}^{2n}
W_i^{(c)}
\left(
\boldsymbol{\chi}_i(t_1)-\mathbf{x}_2(t_1)
\right)
\left(
\boldsymbol{\chi}_i(t_1)-\mathbf{x}_2(t_1)
\right)^T 
\end{equation}

The resulting covariance matrix $\mathbf{P}_2(t_1)$ is  used in the sensitivity analysis step by comparing it with the original CDM covariance matrix $\mathbf{P}_1(t_1)$. The same procedure is repeated for all parameters included in the $\boldsymbol{\theta}$ vector. Thus, the contribution of each parameter to the covariance change is quantified.

\subsection{Sensitivity Analysis}
Sensitivity analysis is performed to quantify how variations in individual parameters of the $\boldsymbol{\theta}$ vector affect the propagation of uncertainty.\cite{ref14} Since the orbital dynamics are nonlinear, even small variations applied to a single parameter may lead to noticeable changes in the propagated covariance matrix. Therefore, a systematic metric is required to measure the overall difference between covariance matrices obtained under different parameter perturbations. The first matrix represents the baseline case, where the system is propagated without any additional controlled variation. The second matrix is obtained by introducing a controlled variation to a single parameter in the $\boldsymbol{\theta}$ vector and propagating the system again through the UT based  backpropagation and propagation process. $\theta_i$ denotes the $i$-th parameter of the parameter vector $\boldsymbol{\theta}$. A controlled  variation is applied to this parameter as
\begin{equation}
\theta_i' = \theta_i + \Delta \theta_i 
\label{eq:theta_variation}
\end{equation}
The relative change in the input parameter is defined as
\begin{equation}
\varepsilon_{\theta_i} =
\frac{\Delta \theta_i}{\theta_i}
\label{eq:relative_input_change}
\end{equation}
The Frobenius norm is used as the primary metric to measure the magnitude of the covariance matrix 
variation. When comparing two matrices, the Frobenius norm is applied to their difference in order to quantify 
the overall variation:
\begin{equation}
\left\| \mathbf{A} - \mathbf{B} \right\|_F =
\sqrt{
\sum_{i=1}^{n}
\sum_{j=1}^{n}
\left(a_{ij}-b_{ij}\right)^2
}
\label{eq:frobenius_difference}
\end{equation}
In the proposed workflow, the baseline covariance matrix at the original CDM epoch is denoted by $\mathbf{P}_{base}(t_1)$, and the covariance matrix obtained after perturbing the $i$-th parameter is denoted by $\mathbf{P}_{pert}^{(i)}(t_1)$. The covariance variation is defined as
\begin{equation}
\Delta \mathbf{P}^{(i)} =
\mathbf{P}_{pert}^{(i)}(t_1) -
\mathbf{P}_{base}(t_1)
\label{eq:covariance_difference}
\end{equation}
To quantify the relative magnitude of this variation, the percentage covariance change based on the Frobenius norm is computed as
\begin{equation}
E_F^{(i)}
=
100 \times
\frac{
\left\|
\mathbf{P}_{pert}^{(i)}(t_1) -
\mathbf{P}_{base}(t_1)
\right\|_F
}{
\left\|
\mathbf{P}_{base}(t_1)
\right\|_F
}
\label{eq:frobenius_sensitivity_metric}
\end{equation}
$E_F^{(i)}$ represents the percentage change in the covariance matrix caused by the variation of the $i$-th parameter. By using this metric, the relationship between the input parameter variation and the resulting covariance variation can be directly evaluated. The dataset generation workflow follows a one-at-a-time parameter variation strategy within the UT propagation loop. At each step, only one parameter is perturbed while all remaining parameters are held 
constant. The perturbed parameters used in the UT workflow are listed in Table~\ref{tab:perturbed_parameters}.
\begin{table}[h!]
\centering
\caption{Perturbed Parameters of The $\boldsymbol{\theta}$ Vector in The UT Workflow.}
\label{tab:perturbed_parameters}
\begin{tabular}{c|c}
\hline
\textbf{Perturbed Parameter Vector} & \textbf{Variation Unit} \\
\hline
$x+\Delta x,\, y,\, z,\, v_x,\, v_y,\, v_z,\, C_d,\, C_r,\, A$ & km \\
$x,\, y+\Delta y,\, z,\, v_x,\, v_y,\, v_z,\, C_d,\, C_r,\, A$ & km \\
$x,\, y,\, z+\Delta z,\, v_x,\, v_y,\, v_z,\, C_d,\, C_r,\, A$ & km \\
$x,\, y,\, z,\, v_x+\Delta v_x,\, v_y,\, v_z,\, C_d,\, C_r,\, A$ & km/s \\
$x,\, y,\, z,\, v_x,\, v_y+\Delta v_y,\, v_z,\, C_d,\, C_r,\, A$ & km/s \\
$x,\, y,\, z,\, v_x,\, v_y,\, v_z+\Delta v_z,\, C_d,\, C_r,\, A$ & km/s \\
$x,\, y,\, z,\, v_x,\, v_y,\, v_z,\, C_d+\Delta C_d,\, C_r,\, A$ & unitless \\
$x,\, y,\, z,\, v_x,\, v_y,\, v_z,\, C_d,\, C_r+\Delta C_r,\, A$ & unitless \\
$x,\, y,\, z,\, v_x,\, v_y,\, v_z,\, C_d,\, C_r,\, A+\Delta A$ & m$^2$ \\
\hline
\end{tabular}
\end{table}

The covariance variation obtained by applying a separate $1\%$ increase to each parameter is presented in Table~\ref{tab:sensitivity_results}. The results indicate that even relatively small variations in the input parameters lead to significant changes in the propagated covariance matrix. Although the applied input variation is only $1\%$, the resulting covariance variation reaches considerably higher levels.

\begin{table}[h!]
\centering
\caption{Sensitivity Analysis Results.}
\label{tab:sensitivity_results}
\begin{tabular}{c|c|c}
\hline
\textbf{Input Parameter} & \textbf{Input Variation (\%)} & \textbf{Covariance Matrix Change (\%)} \\
\hline
Position $x$ & 1 & 38.409748 \\
Position $y$ & 1 & 38.937861 \\
Position $z$ & 1 & 36.916528 \\
Velocity $v_x$ & 1 & 67.838004 \\
Velocity $v_y$ & 1 & 66.341729 \\
Velocity $v_z$ & 1 & 66.929911 \\
Drag Coefficient $(C_d)$ & 1 & 39.448682 \\
SRP Coefficient $(C_r)$ & 1 & 30.173464 \\
Cross-Sectional Area $(A)$ & 1 & 31.010773 \\
\hline
\end{tabular}
\end{table}

\subsection{Principal Component Analysis}
\begin{figure}[!htbp]
    \centering
    \includegraphics[width=3.8in]{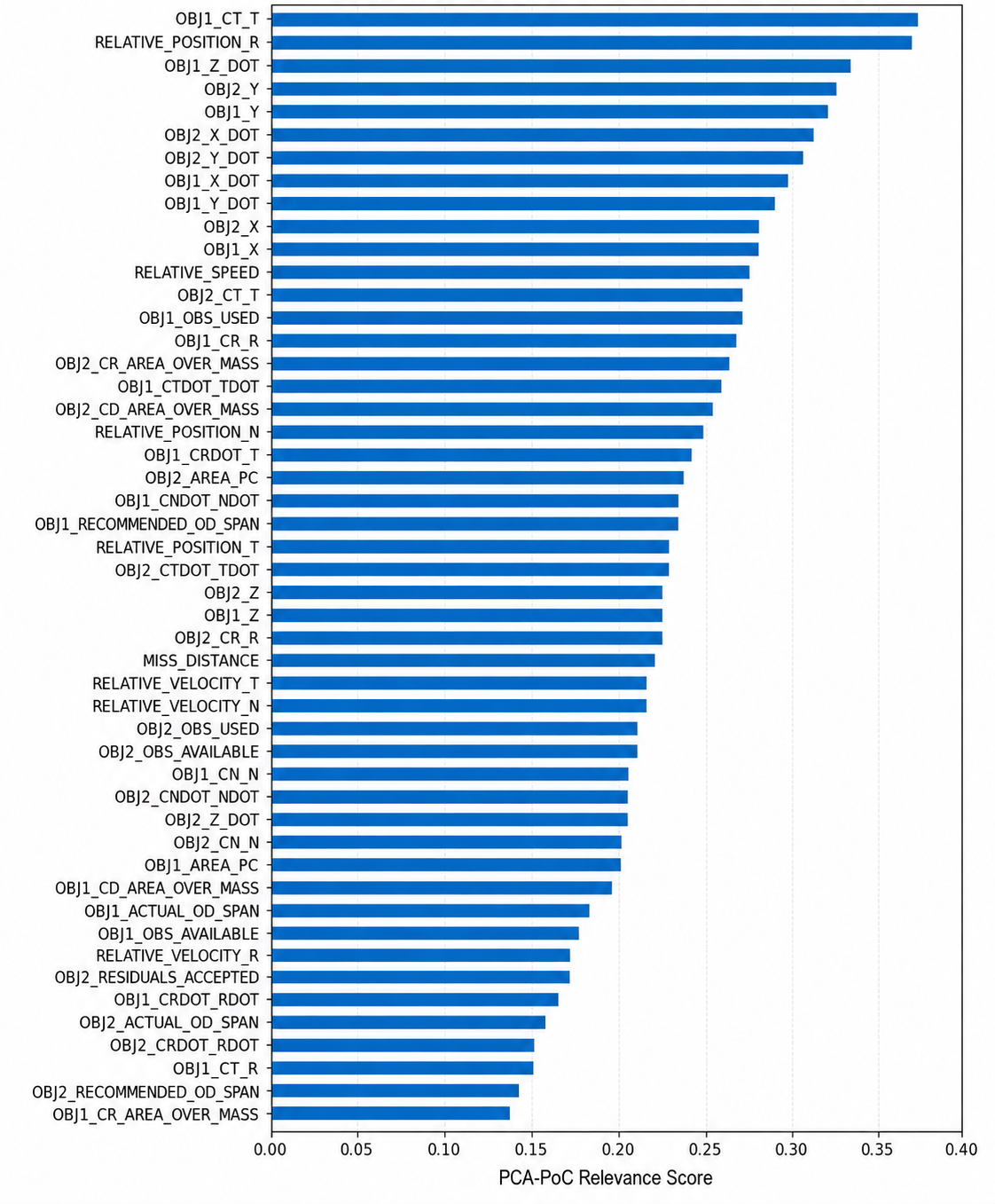}
    \caption{Raw CDM parameters ranked according to the PCA--PoC relevance score.}
    \label{fig:pcaMain}
\end{figure}

PCA is used to identify the numerical CDM parameters that are most relevant to the probability of collision. In this study, PCA is applied using the built in $\texttt{pca}$ function in MATLAB, which transforms the standardized CDM parameters into orthogonal principal components representing the dominant variance directions in the dataset. The PCA-PoC relevance score is computed by combining the PCA loadings, explained variance ratios, and the correlation between principal component scores and PoC values.\cite{ref15} This score does not represent a direct parameter--PoC correlation; instead, it indicates how strongly each CDM parameter contributes to variance directions associated with PoC variation. Figure~\ref{fig:pcaMain} presents the raw numerical CDM parameters ranked according to their PCA--PoC relevance scores. These results are used to determine the most relevant parameters to be included in the proposed dataset.
 
\subsection{Dataset Structure}
The dataset structure is defined according to the findings obtained from the sensitivity analyses and PCA. The sensitivity results show that small perturbations in the state vector and physical parameters used in orbit propagation can significantly affect the propagated covariance structure. Since the covariance represents the uncertainty of the space object at the encounter epoch, these uncertainties influence the resulting probability of collision as shown in Eq.~\ref{eq:POC}. Table~\ref{tab:rawDataset} presents the raw parameters extracted from each CDM and included in the dataset. Since the covariance represents the uncertainty of the space object at the encounter epoch, these uncertainties influence the resulting probability of collision as shown in Eq.~\ref{eq:POC}. Table~\ref{tab:rawDataset} presents the raw parameters extracted from each CDM and included in the dataset. Similarly, the PCA results indicate that the parameters carrying the dominant information related to PoC are mainly associated with relative position, relative velocity, covariance terms, close approach time, and physical coefficients. Therefore, both raw CDM parameters and derived physical metrics from selected parameters are included in the dataset. The raw CDM parameters describe the relative state and uncertainty representation of the objects, while the derived metrics provide additional information on the geometry and dynamical characteristics of the conjunction.

\begin{table}[htbp]
\centering
\caption{Raw CDM Parameters in The Dataset.}
\label{tab:rawDataset}
\begin{tabular}{lc}
\hline
\textbf{Raw CDM Parameter} & \textbf{Unit} \\
\hline
Event ID & unitless \\
CDM Creation Date [UTC] & unitless \\
TCA [UTC] & unitless \\
Relative Position Norm & km \\
Position $R, T, N$ & km \\
Velocity $R, T, N$ & km/s \\
Relative Speed & km/s \\
Miss Distance & km \\
Probability of Collision & unitless \\
State Vector for Object 1 and Object 2 & km, km/s \\
$C_d \cdot$ Area/Mass for Object 1 and Object 2 & m$^2$/kg \\
$C_r \cdot$ Area/Mass for Object 1 and Object 2 & m$^2$/kg \\
Area/Mass for Object 1 and Object 2 & m$^2$/kg \\
\hline
\end{tabular}
\end{table}

Additional physically meaningful metrics are derived from the raw state, covariance, and probability information to represent the conjunction event more comprehensively. These metrics capture the encounter geometry, uncertainty distribution, relative motion characteristics, and temporal evolution of the event, which are not explicitly provided by the raw CDM parameters. The derived metrics included in the dataset are listed in Table~\ref{tab:derived_metrics}. The learning model is provided with more informative physical representations of the conjunction event by incorporating these parameters into the training dataset. This enhanced feature representation contributes to improving the accuracy of the predicted probability of collision.

\begin{table}[htbp]
\centering
\caption{Derived Physical Metrics in The Dataset.}
\label{tab:derived_metrics}
\begin{tabular}{lc}
\hline
\textbf{Derived Physical Metric} & \textbf{Unit} \\
\hline
Time difference between CDM Creation Date and TCA & day \\
2D Collision Plane Distance & km \\
Standard Deviations of Position Covariance & km \\
3D Covariance Ellipsoid Volume & km$^3$ \\
Mahalanobis Distance & unitless \\
Relative Motion Geometry Approach Angle & deg \\
Relative Motion Geometry OOP Ratio & unitless \\
Logarithmic Scale of Probability of Collision & unitless \\
Trend PoC & day$^{-1}$ \\
Trend Miss Distance & km/day \\
Trend Relative Position & km/day \\
Trend Relative Speed & km/s/day \\
Classification Label & unitless \\
\hline
\end{tabular}
\end{table}

\subsubsection{2D Collision Plane Distance}
The 2D collision plane distance is defined as the magnitude of the relative position vector projected onto the encounter plane. The encounter plane is perpendicular to the relative velocity vector at TCA. Therefore, this distance represents the separation between the two objects in the plane where the collision probability is evaluated. At TCA, the conjunction geometry is described using the relative position and velocity vectors, $\mathbf{r}_{rel}$ and $\mathbf{v}_{rel}$. The unit vector along the relative velocity, the parallel projection, and the perpendicular component are defined as
\begin{equation}
\hat{\mathbf{e}}_{v} = \frac{\mathbf{v}_{rel}}{\|\mathbf{v}_{rel}\|}, 
\qquad
r_{\parallel} = \mathbf{r}_{rel}^{T}\hat{\mathbf{e}}_{v},
\qquad
\mathbf{r}_{\perp} = \mathbf{r}_{rel} - r_{\parallel}\hat{\mathbf{e}}_{v}
\end{equation}

The 2D collision plane distance is obtained as
\begin{equation}
d_{\mathrm{2D}} = \|\mathbf{r}_{\perp}\|
\end{equation}

\subsubsection{Standard Deviations of Position Covariance}
The standard deviations $\sigma_R$, $\sigma_T$, and $\sigma_N$ represent the uncertainty of the relative position in the radial, along-track, and normal directions of the RTN frame. Assuming independent object uncertainties, the relative covariance is obtained from the covariance matrices of the two objects, and the standard deviations are computed from its diagonal elements as
\begin{equation}
\sigma_R = \sqrt{C_{RR}}, \quad
\sigma_T = \sqrt{C_{TT}}, \quad
\sigma_N = \sqrt{C_{NN}}
\end{equation}

\subsubsection{3D Covariance Ellipsoid Volume}
The 3D covariance ellipsoid volume provides a scalar measure of the overall positional uncertainty of the relative state in the RTN frame as expressed in Eq.~\eqref{eq:volume}.
\begin{equation}
V = \frac{4}{3}\pi \sigma_R \sigma_T \sigma_N 
\label{eq:volume}
\end{equation}
This metric combines the uncertainty contributions in all three RTN directions into a single value. It helps characterize how the uncertainty region may influence the collision risk.

\subsubsection{Mahalanobis Distance}
The Mahalanobis distance expresses the separation between a point and a distribution by considering the covariance structure of the data. Its general form is defined as

\begin{equation}
d_{\mathrm{Mahalanobis}} =
\sqrt{
(\mathbf{x}-\boldsymbol{\mu})^{T}
\mathbf{P}^{-1}
(\mathbf{x}-\boldsymbol{\mu})
}
\end{equation}

where $\mathbf{x}$ is the observed vector, $\boldsymbol{\mu}$ is the mean vector, and $\mathbf{P}$ is the covariance matrix. In conjunction analysis, this metric is evaluated in the encounter plane to express the projected miss distance relative to the covariance-based uncertainty ellipse. The projected miss vector and the corresponding covariance matrix in the encounter plane are defined as
\begin{equation}
\mathbf{d}_{enc} =
\begin{bmatrix}
x_m \\
y_m
\end{bmatrix},
\qquad
\mathbf{C}_{enc} =
\begin{bmatrix}
\sigma_x^2 & C_{xy} \\
C_{xy} & \sigma_y^2
\end{bmatrix}
\end{equation}
The Mahalanobis distance for the encounter plane is then computed as
\begin{equation}
d_{\mathrm{Mahalanobis}} =
\sqrt{
\mathbf{d}_{enc}^{T}
\mathbf{C}_{enc}^{-1}
\mathbf{d}_{enc}
}
\end{equation}
If the covariance coupling term is neglected, i.e., $C_{xy}=0$, the expression reduces to
\begin{equation}
d_{\mathrm{Mahalanobis}} =
\sqrt{
\frac{x_m^2}{\sigma_x^2}
+
\frac{y_m^2}{\sigma_y^2}
}
\end{equation}
When the Mahalanobis distance is smaller than 1, the projected relative position is located inside the $1\sigma$ uncertainty ellipse. When it is equal to 1, the position lies on the ellipse boundary, while values greater than 1 indicate that the projected relative position is outside the $1\sigma$ uncertainty region.\cite{ref16}

\subsubsection{Approach Angle in Encounter Geometry}
The approach angle is a geometric feature used to characterize the relative motion between two space objects at TCA. It is defined as the angle between the relative position vector and the relative velocity vector. This metric indicates whether the encounter geometry is closer to a parallel, perpendicular, or opposite-direction motion.
\begin{equation}
\theta_{\mathrm{approach}} =
\cos^{-1}
\left(
\frac{\mathbf{r}_{rel} \cdot \mathbf{v}_{rel}}
{\|\mathbf{r}_{rel}\| \, \|\mathbf{v}_{rel}\|}
\right)
\end{equation}
An approach angle equal to or close to $0^\circ$ indicates that the relative position and relative velocity vectors are nearly aligned, representing a parallel-motion type encounter. An angle close to $90^\circ$ represents a perpendicular encounter geometry, whereas an angle close to $180^\circ$ indicates opposite-direction motion, corresponding to a head-on type encounter.\cite{ref17}
 
\subsubsection{Out of Plane Ratio}
The out-of-plane (OOP) ratio provides a compact measure of how much the encounter geometry deviates from the orbital plane. The OOP ratio is defined as
\begin{equation}
\alpha =
\frac{|N|}{\sqrt{R^2 + T^2}}
\end{equation}
$R$, $T$, and $N$ denote the radial, along-track, and normal components of the relative position vector. A value of $\alpha$ close to zero indicates that the encounter occurs almost entirely within the orbital plane, whereas a value close to one indicates that the encounter takes place largely outside the orbital plane.\cite{ref17}

\subsubsection{Trend Based Features}
Trend-based features are used to describe the temporal evolution of selected CDM parameters across multiple messages belonging to the same event. Instead of considering only the instantaneous value of a parameter, these features quantify whether the parameter increases or decreases as the conjunction event approaches. In this study, trend values are computed for the probability of collision, miss distance, relative position, and relative speed. The temporal variation of a parameter is modeled using a linear regression formulation as
\begin{equation}
x(t) = \beta t + \gamma 
\end{equation}
$x(t)$ represents the parameter value at time $t$, $\beta$ is the slope corresponding to the trend, and $\gamma$ is the intercept. The slope is calculated using the least-squares method as
\begin{equation}
\beta =
\frac{
\sum_{i=1}^{n}(t_i-\bar{t})(x_i-\bar{x})
}{
\sum_{i=1}^{n}(t_i-\bar{t})^2
}
\end{equation}
 $x_i$ denotes the value of the analyzed parameter at the corresponding CDM time $t_i$. A positive slope indicates an increasing trend, while a negative slope indicates a decreasing trend.\cite{ref18}

\subsubsection{Classification Label}
In addition to the continuous PoC value used for regression, a binary risk label is generated to support the classification task. The label is assigned according to a predefined PoC threshold, which separates low-risk and high-risk conjunction cases. In this study, the threshold is selected as $10^{-6}$, and the risk label is defined as
\begin{equation}
\mathrm{Risk\ Label} =
\begin{cases}
1, & P_c \geq 10^{-6} \\
0, & P_c < 10^{-6}
\end{cases}
\label{eq:risk_label}
\end{equation}
A label of 1 represents a potentially risky conjunction event, while a label of 0 represents a low-risk case. This formulation enables the dataset to be used not only for PoC prediction as a regression problem, but also for identifying critical conjunction conditions through binary classification.

Following the extraction of raw CDM parameters and the computation of derived physical metrics from CDM, each row of the final dataset is defined as an enriched conjunction data. It does not represent only the original CDM; instead, it combines the raw parameters extracted from the CDM with the additional physical metrics derived from state, covariance, geometry, and temporal information. The $i$-th enriched conjunction data is denoted by $\mathbf{x}_i$ and defined as
\begin{equation}
\mathbf{x}_i =
\left[
\mathbf{x}_{i}^{raw},
\mathbf{x}_{i}^{derived}
\right]
\end{equation}
where $\mathbf{x}_{i}^{raw}$ represents the raw CDM parameter vector and $\mathbf{x}_{i}^{derived}$ represents the derived physical metric vector. In the constructed dataset, the raw CDM parameter vector contains 31 features, while the derived physical metric vector contains 13 features:
\begin{equation}
\mathbf{x}_{i}^{raw} \in \mathbb{R}^{31},
\qquad
\mathbf{x}_{i}^{derived} \in \mathbb{R}^{13}
\end{equation}
Each enriched conjunction data  is represented by a 44-dimensional input feature vector:
\begin{equation}
\mathbf{x}_i \in \mathbb{R}^{44}
\end{equation}
If the dataset contains $N$ enriched conjunction data samples, the complete input matrix is expressed as
\begin{equation}
\mathbf{X}
=
\begin{bmatrix}
\mathbf{x}_1^{T} \\
\mathbf{x}_2^{T} \\
\vdots \\
\mathbf{x}_N^{T}
\end{bmatrix}
\in \mathbb{R}^{N \times 44}
\end{equation}

\section{Neural Network Framework}
The objective of this study is to predict the temporal evolution of collision risk during the early stages of a conjunction event using the created dataset. For this purpose, a hybrid TCN-Transformer model is employed to estimate the PoC value of a future CDM associated with the same close approach event. This early PoC estimation is intended to assist satellite operations by providing risk information before the next CDM update becomes available.

\subsection{Mathematical Background of Hybrid TCN-Transformer Model}
The constructed dataset was used to train a hybrid TCN–Transformer neural network to model the evolution of conjunction risk. This architectural choice is driven by the sequential nature of CDM, which is organized as Event ID–based time series. The TCN sub-network captures short-term temporal variations between consecutive CDMs, enabling the estimation of risk behavior from limited prior information. Even with only one or two CDMs, the TCN learns local transitions in $P_c$, miss distance, and state parameters, supporting early-stage risk assessment.~\cite{ref19} In contrast, when longer CDM sequences are available, the Transformer sub-network captures long range dependencies across the entire sequence.~\cite{ref20}  By modeling how earlier covariance and geometric changes influence later updates, it learns event level risk evolution and anticipates subsequent CDM behavior. For a given conjunction event $e$, the enriched conjunction data samples belonging to the same Event ID are ordered chronologically according to their CDM creation time. The sequential input of event $e$ is defined as
\begin{equation}
\mathbf{X}_{e}
=
\left[
\mathbf{x}_{1},
\mathbf{x}_{2},
\ldots,
\mathbf{x}_{k}
\right]
\end{equation}
where $\mathbf{X}_{e}$ denotes the enriched conjunction data sequence of event $e$, $\mathbf{x}_{t}$ is the $t$-th enriched conjunction data sample in this event sequence, and $k$ is the number of available enriched conjunction data samples used as input.
\begin{equation}
\mathbf{X}_{e}
=
\begin{bmatrix}
\mathbf{x}_{1}^{T} \\
\mathbf{x}_{2}^{T} \\
\vdots \\
\mathbf{x}_{k}^{T}
\end{bmatrix}
\in \mathbb{R}^{k \times 44}
\end{equation}
The proposed model directly predicts the probability of collision value. The objective is to estimate the PoC value of the future CDM update using the available previous CDM information. For example, using the first three CDM updates, the model predicts the PoC value of the fourth CDM update:
\begin{equation}
\left[
CDM_{1}, CDM_{2}, CDM_{3}
\right]
\rightarrow
\widehat{PoC}_{4}
\end{equation}
In the general form, the prediction problem is expressed as
\begin{equation}
PoC_{k+1}^{(t)}
=
f_{\theta^{(t)}}
\left(
\mathbf{x}_{1},
\mathbf{x}_{2},
\ldots,
\mathbf{x}_{k}
\right)
\end{equation}

where $PoC_{k+1}^{(t)}$ is the predicted PoC value for the next CDM, 
$f_{\theta^{(t)}}$ denotes the hybrid TCN--Transformer model with learned 
parameters $\theta^{(t)}$, and $\mathbf{x}_{i}$ is the feature vector of the 
$i$th observed data.

\subsubsection{Temporal Convolutional Network}
The enriched conjunction data sequence is given to the TCN layer. The main purpose of
the TCN is to learn the temporal variation between consecutive data updates. The TCN captures how the conjunction-related parameters change over time, such as whether the PoC increases or decreases, whether the miss distance becomes smaller or larger, how the relative state vary, how the covariance-based uncertainty evolves, and how the Mahalanobis distance changes.\cite{ref19} The output of the TCN is written as
\begin{equation}
\mathbf{H}_{\mathrm{TCN}}
=
\mathrm{TCN}
\left(
\mathbf{X}_{e}
\right)
\end{equation}
\begin{equation}
\mathbf{H}_{\mathrm{TCN}}
=
\left[
\mathbf{h}_{1},
\mathbf{h}_{2},
\ldots,
\mathbf{h}_{k}
\right]
\end{equation}
Each $\mathbf{h}_{i}$ is the enriched temporal representation of the corresponding data update after the TCN operation. The TCN applies convolution along the temporal axis. The basic dilated causal convolution operation can be written as
\begin{equation}
h_t = \mathrm{ReLU}\left(\sum_{k=0}^{K-1} w_k \cdot x(t - d \cdot k) + b \right)
\end{equation}
where $\mathbf{h}_{t}$ is the TCN output at time step $t$, $x(t-d\cdot k)$ denotes the input from the previous time steps, $w_k$ represents the learned filter weight, $K$ is the kernel size, $d$ is the dilation factor, and $b$ is the bias term and $\mathrm{ReLU}$ is the activation function.\cite{ref19} The dilation factor allows the TCN to observe not only the immediately previous data update but also earlier enriched conjunction data updates over a wider temporal history. Therefore, the TCN extracts local temporal evolution information between conjunction data updates. The activation function used in the TCN layers is the rectified linear unit function. It is defined as
\begin{equation}
\mathrm{ReLU}(z)
=
\max(0,z)
\end{equation}
If the input value is negative, the output becomes zero. If the input value is positive, it passes through unchanged. This activation function enables the model to learn nonlinear relationships between the input features. In this problem, the relationship between PoC, miss distance, covariance, and relative motion parameters is not necessarily linear. Therefore, the ReLU activation function allows the TCN to learn nonlinear temporal patterns in the CDM sequence.

\subsubsection{Transformer}
After the TCN block, the obtained temporal feature sequence, $\mathbf{H}_{\mathrm{TCN}}$, is given to the Transformer encoder. The task of the Transformer is to determine which enriched conjunction data update in the sequence is more important for the PoC prediction.  An early CDM can have high uncertainty, while a later CDM can provide more reliable information. The Transformer does not inherently know the temporal order of the sequence. Therefore, positional encoding is added to the TCN output in order to preserve the chronological order of the CDM updates:
\begin{equation}
\mathbf{Z}
=
\mathbf{H}_{\mathrm{TCN}}
+
\mathbf{P}_{\mathrm{pos}}
\end{equation}
where $\mathbf{H}_{\mathrm{TCN}}$ is the TCN output, $\mathbf{P}_{\mathrm{pos}}$ is the positional encoding matrix, and $\mathbf{Z}$ is the input of the Transformer encoder. This operation allows the model to distinguish the order of the CDM related data updates such as $CDM_{1}$, $CDM_{2}$, and $CDM_{3}$. The Transformer learns these relationships using the self-attention mechanism. In the Transformer encoder, three matrices are  computed:
\begin{equation}
\mathbf{Q}
=
\mathbf{Z}\mathbf{W}_{Q},
\qquad
\mathbf{K}
=
\mathbf{Z}\mathbf{W}_{K},
\qquad
\mathbf{V}
=
\mathbf{Z}\mathbf{W}_{V}
\end{equation}
where $\mathbf{Q}$ is the query matrix, $\mathbf{K}$ is the key matrix, $\mathbf{V}$ is the value matrix, and $\mathbf{W}_{Q}$, $\mathbf{W}_{K}$, and $\mathbf{W}_{V}$ are learnable weight matrices. The attention operation is calculated by using softmax activation function as
\begin{equation}
\mathrm{Attention}
\left(
\mathbf{Q},
\mathbf{K},
\mathbf{V}
\right)
=
\mathrm{softmax}
\left(
\frac{
\mathbf{Q}
\mathbf{K}^{T}
}{
\sqrt{d_{k}}
}
\right)
\mathbf{V}
\end{equation}
The softmax function normalizes the attention scores and converts them into importance weights:
\begin{equation}
\mathrm{softmax}(a_i)
=
\frac{
e^{a_i}
}{
\sum_j e^{a_j}
}
\end{equation}
The sum of all attention weights becomes equal to one, and the model assigns an importance weight to each CDM update. This operation calculates the relationship between each CDM update and the other data updates in the sequence. As a result, the model learns which CDM updates are more important for predicting the PoC value. The second activation function is the ReLU function used in the feed-forward network of the Transformer. The feed-forward network is expressed as
\begin{equation}
\mathrm{FFN}
\left(
\mathbf{z}
\right)
=
\mathbf{W}_{2}
\mathrm{ReLU}
\left(
\mathbf{W}_{1}
\mathbf{z}
+
\mathbf{b}_{1}
\right)
+
\mathbf{b}_{2}
\end{equation}
The ReLU function provides nonlinear transformation in the feed-forward part of the Transformer. The self-attention output and the feed-forward output are combined through residual connections and layer normalization within each Transformer encoder block. The output of the Transformer encoder is written as
\begin{equation}
\mathbf{Z}_{\mathrm{transformer}}
=
\mathrm{Transformer}
\left(
\mathbf{H}_{\mathrm{TCN}}
\right)
\end{equation}
Here, $\mathbf{z}_{1}$ is the attention-based representation of the first CDM update, $\mathbf{z}_{k}$ is the attention-based representation of the last CDM update, and $\mathbf{Z}_{\mathrm{enc}}$ is the complete sequence representation produced by the Transformer. This output contains both the local temporal information extracted by the TCN and the global sequence-level relationships learned by the Transformer.

\subsubsection{PoC regression and model training}
The Transformer encoder output is obtained as a sequence of hidden representations:
\begin{equation}
\mathbf{Z}_{\mathrm{transformer}}
=
\left[
\mathbf{z}_{1},
\mathbf{z}_{2},
\ldots,
\mathbf{z}_{k}
\right]
\end{equation}

Since the objective is to predict the PoC value of the next CDM update using the available event history, the hidden representation of the last available enriched conjunction data sample is used as the event-level representation:
\begin{equation}
\mathbf{z}_{e}
=
\mathbf{z}_{k}
\end{equation}
The predicted PoC value is then obtained through a linear regression layer:
\begin{equation}
\hat{Pc}_{k+1}^{(e)}
=
\mathbf{W}_{o}\mathbf{z}_{e}
+
b_{o}
\end{equation}
where $\mathbf{W}_{o}$ and $b_{o}$ are the trainable parameters of the output layer. Since PoC prediction is formulated as a regression problem, a linear activation function is used at the output layer:
\begin{equation}
g(z)=z
\end{equation}
The model is trained by minimizing the Smooth L1 loss between the true and predicted PoC values. For $S$ training sequence-target pairs, the loss function is defined as
\begin{equation}
\mathcal{L}_{\mathrm{SmoothL1}} =
\frac{1}{S}
\sum_{s=1}^{S}
\begin{cases}
\frac{1}{2}\left(P_{c,s}-\hat{P}_{c,s}\right)^2, 
& \left|P_{c,s}-\hat{P}_{c,s}\right| < 1, \\[4pt]
\left|P_{c,s}-\hat{P}_{c,s}\right|-\frac{1}{2}, 
& \left|P_{c,s}-\hat{P}_{c,s}\right| \geq 1.
\end{cases}
\label{eq:loss}
\end{equation}
where $P_{c,s}$ is the true PoC value and $\hat{P}_{c,s}$ is the predicted PoC value for the $s$-th training sequence. 
The optimal model parameters are obtained by solving
\begin{equation}
\boldsymbol{\theta}^{*}
=
\arg\min_{\boldsymbol{\theta}}
\mathcal{L}_{\mathrm{SmoothL1}}
\end{equation}
During training, the trainable parameters of the TCN block, Transformer encoder, and regression head are updated through backpropagation as
\begin{equation}
\boldsymbol{\theta}
\leftarrow
\boldsymbol{\theta}
-
\eta
\nabla_{\boldsymbol{\theta}}
\mathcal{L}_{\mathrm{SmoothL1}}
\end{equation}
where $\eta$ is the learning rate. Through this optimization process, the model learns to map the sequential enriched conjunction data representation of a conjunction event to the predicted PoC value of the next CDM update.

\subsubsection{Dataset partitioning}
 The dataset was divided into training and test sets using an Event ID based separation strategy to ensure realistic performance evaluation. All CDMs belonging to a given Event ID were kept within the same subset. The dataset consists exclusively of real CDMs associated with LEO satellites. The resulting event-based dataset partitioning is summarized in Table~\ref{tab:dataset_split}.
\begin{table}[htbp]
\centering
\caption{Event Based Dataset Partitioning.}
\label{tab:dataset_split}
\begin{tabular}{lcc}
\hline
\textbf{Subset} & \textbf{Number of Events} & \textbf{Percentage} \\
\hline
Training   & 856  & 79.63\% \\
Validation & 86   & 8.00\%  \\
Test       & 133  & 12.37\% \\
\hline
Total      & 1075 & 100.00\% \\
\hline
\end{tabular}
\end{table}

\subsubsection{Model hyperparameters}
The training-related parameters are configured to improve the stability and predictive performance of the model. The selected hyperparameters used in the training process are summarized in Table~\ref{tab:hybridHyperparameters}.
\begin{table}[htbp]
\centering
\caption{Hybrid TCN--Transformer Training Hyperparameters.}
\label{tab:hybridHyperparameters}
\begin{tabular}{lc}
\hline
\textbf{Parameter} & \textbf{Value} \\
\hline
Threshold & $1 \times 10^{-6}$ \\
Seed & 42 \\
Epochs & 200 \\
Learning Rate & $3 \times 10^{-4}$ \\
Weight Decay & $1 \times 10^{-4}$ \\
Training Batch Size & 32 \\
Evaluation Batch Size & 64 \\
\hline
TCN Channels & $[64, 64, 64, 64]$ \\
TCN Kernel Size & 5 \\
TCN Dropout & 0.03 \\
\hline
Model Dimension & 64 \\
Attention Heads & 4 \\
Transformer Layers & 2 \\
Feedforward Multiplier & 4 \\
Attention Dropout & 0.05 \\
Max Sequence Length & 256 \\
\hline
\end{tabular}
\end{table}

\subsubsection{Prediction Performance.}
Model hyperparameters are selected based on empirical validation and commonly adopted practices in time-series learning. The model is trained for 200 epochs using the AdamW optimizer with a learning rate of $3\times10^{-4}$ and a weight decay coefficient of $10^{-4}$ to reduce overfitting and improve generalization performance on unseen conjunction events.~\cite{ref19, ref20} A Smooth L1 loss function is adopted to improve training stability and reduce sensitivity to relatively large prediction deviations.~\cite{ref21}

\section{Results}
\subsection{Learning Behavior of the Proposed Hybrid Model}
\begin{figure}[htbp]
\centering
    \centering\includegraphics[width=3.5in]{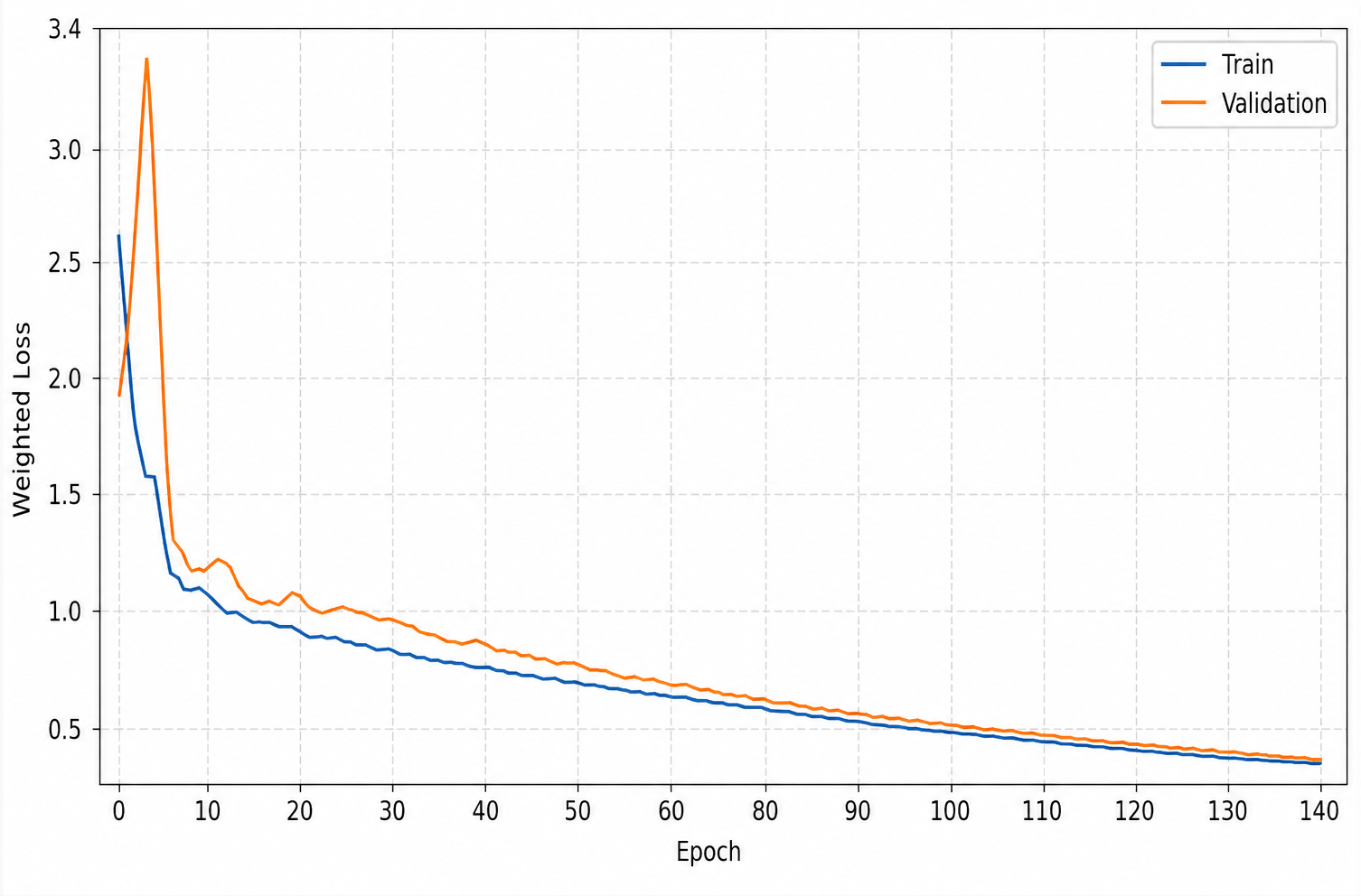}
    \caption{Training and Validation Loss Curves of the Hybrid TCN--Transformer Model}
    \label{fig:HybridTrainingCurve}
\end{figure}

Figure~\ref{fig:HybridTrainingCurve} presents the training and validation loss curves of the proposed hybrid TCN--Transformer model. The curves represent the evolution of the Smooth L1 loss function defined in Eq.~\eqref{eq:loss} during the training process. Both curves exhibit a consistent decreasing trend, indicating that the model progressively learns the relationship between the sequential enriched conjunction data inputs and the target PoC values. The validation loss follows the training loss closely throughout the training process, which suggests stable convergence without significant overfitting. This behavior demonstrates that the proposed hybrid architecture effectively learns both local temporal variations through the TCN component and sequence-level dependencies through the Transformer encoder.

\subsubsection{Future PoC Prediction}
The trained model is evaluated on the test dataset to assess its next-step PoC prediction performance. Table~\ref{tab:poc_results} presents selected close approach events from the test set, where the predicted PoC values are compared with the corresponding true PoC values obtained from the CDM.
\begin{table}[htbp]
\centering
\caption{Sample Event ID Based Next Step PoC Prediction Results for the hybrid model.}
\label{tab:poc_results}
\begin{tabular}{ccccc}
\hline
\textbf{Event ID} & \textbf{Total CDMs} & \textbf{Observed CDMs} & \textbf{True PoC} & \textbf{Predicted PoC} \\
\hline
81  & 2  & 1  & $5.24 \times 10^{-5}$ & $2.16 \times 10^{-5}$ \\
94  & 5  & 4  & $1.22 \times 10^{-3}$ & $1.96 \times 10^{-4}$ \\
101 & 6  & 5  & $0$                    & $0$ \\
129 & 3  & 2  & $4.43 \times 10^{-4}$ & $7.66 \times 10^{-4}$ \\
142 & 6  & 5  & $1.23 \times 10^{-4}$ & $1.95 \times 10^{-5}$ \\
292 & 2  & 1  & $3.51 \times 10^{-7}$ & $4.19 \times 10^{-7}$ \\
340 & 14 & 13 & $1.29 \times 10^{-5}$ & $4.71 \times 10^{-6}$ \\
362 & 2  & 1  & $1.11 \times 10^{-6}$ & $1.09 \times 10^{-6}$ \\
378 & 5  & 4  & $2.33 \times 10^{-6}$ & $3.45 \times 10^{-6}$ \\
577 & 2  & 1  & $7.38 \times 10^{-8}$ & $6.80 \times 10^{-8}$ \\
586 & 3  & 2  & $7.85 \times 10^{-5}$ & $7.99 \times 10^{-5}$ \\
666 & 3  & 2  & $2.51 \times 10^{-7}$ & $2.20 \times 10^{-7}$ \\
675 & 5  & 4  & $2.37 \times 10^{-6}$ & $1.01 \times 10^{-6}$ \\
777 & 9  & 8  & $3.08 \times 10^{-6}$ & $4.43 \times 10^{-6}$ \\
782 & 6  & 5  & $0$                    & $0$ \\
\hline
\end{tabular}
\end{table}

In this evaluation, each Event ID represents a conjunction event, and the model uses the observed enriched conjunction data sequence to estimate the PoC value of the subsequent CDM update. The Total CDMs column denotes the total number of CDMs available for the corresponding event, while Observed CDMs indicates the number of CDMs used as the model input. The model predicts the PoC value associated with the subsequent CDM update, allowing the temporal evolution of the collision risk to be assessed before the next CDM becomes available. The results show that the proposed hybrid model is able to reproduce the general magnitude and trend of the true PoC values over a wide numerical range, from high-risk cases on the order of $10^{-3}$--$10^{-4}$ to very low-risk cases on the order of $10^{-7}$--$10^{-8}$. For several events, such as Event IDs 292, 362, 577, 586, and 666, the predicted PoC values are very close to the corresponding true values, indicating that the model can capture the temporal risk behavior when the enriched conjunction data sequence contains sufficient information about the risk evolution. For some higher-risk events, such as Event IDs 94 and 142, the model slightly underestimates the true PoC value; however, the predictions remain within the same risk regime and preserve the overall order of magnitude of the collision probability. This indicates that the model is more reliable in capturing the risk level and relative severity of the event than producing an exact point-wise PoC value for every case. Moreover, the zero-PoC events are correctly predicted as zero, showing that the model does not introduce artificial collision risk for events with no estimated probability of collision. Table~\ref{tab:event_classification_metrics} presents the event-based classification performance of the hybrid model. 
\begin{table}[htbp]
\centering
\caption{Event Based Classification Performance Metrics for The Hybrid Model.}
\label{tab:event_classification_metrics}
\begin{tabular}{c c c c c c}
\hline
\textbf{Accuracy} & \textbf{Precision} & \textbf{Recall} & \textbf{Specificity} & \textbf{F1-score} & \textbf{F2-score} \\
\hline
0.9398 & 0.7222 & 0.8125 & 0.9572 & 0.7647 & 0.7926 \\
\hline
\end{tabular}
\end{table}

\begin{figure}[htbp]
    \centering
    \subfigure[Event Based Confusion Matrix]{
        \includegraphics[width=0.48\textwidth]{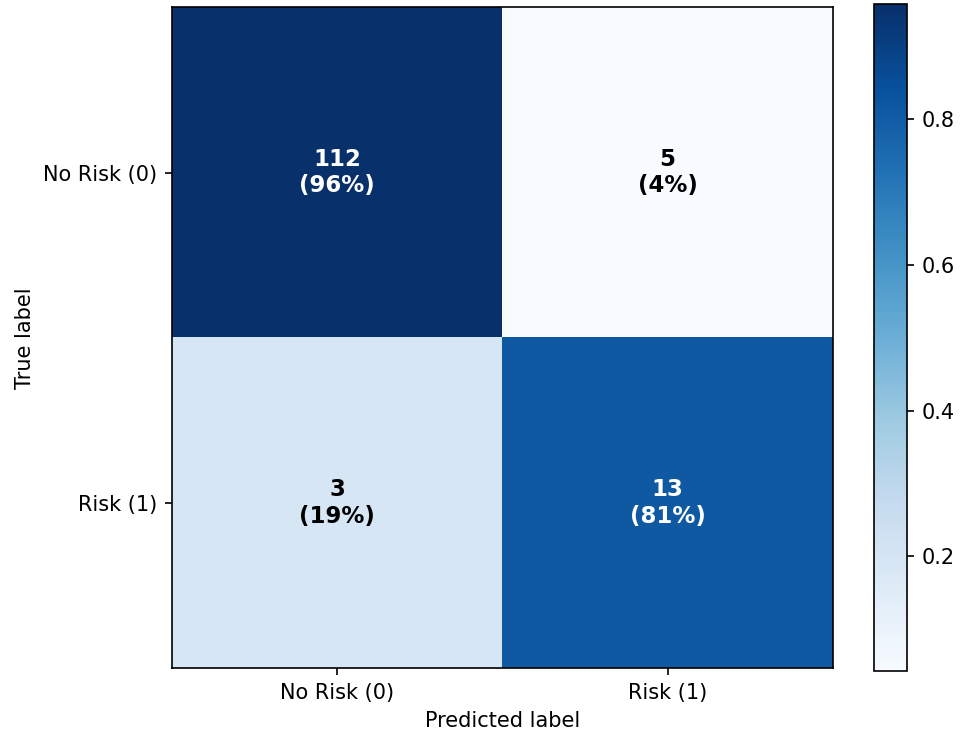}
        \label{fig:ConfusionMatrix}
    }
    \hfill 
    \subfigure[Conjunction Data Based Confusion Matrix]{
        \includegraphics[width=0.48\textwidth]{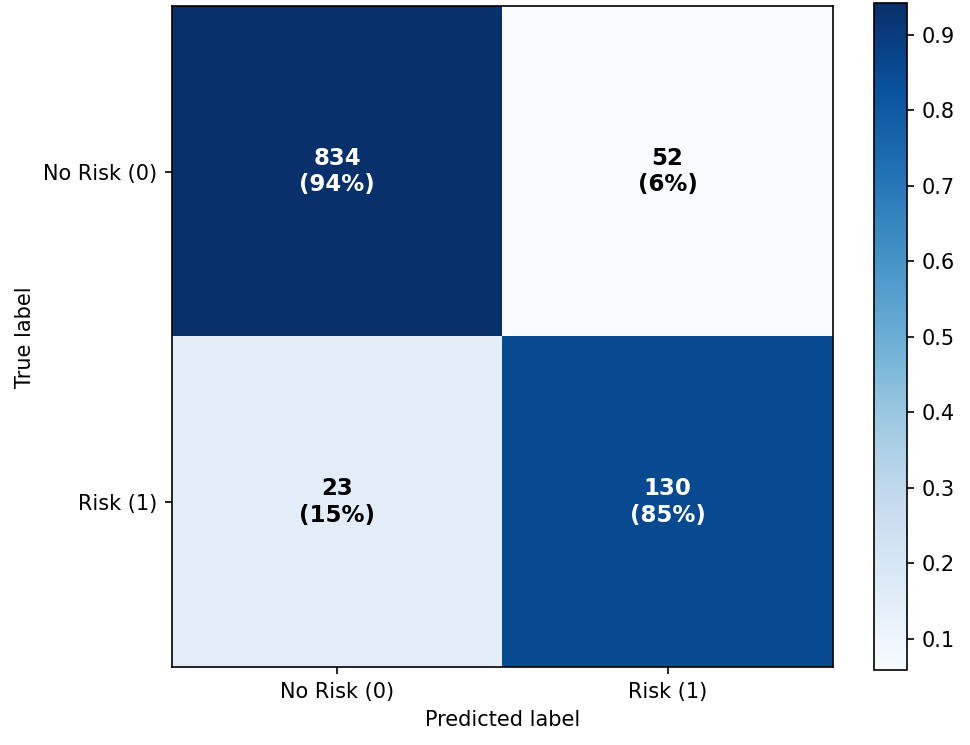}
        \label{fig:ConfusionMatrixPrefix}
    }
    \caption{Confusion Matrices of The Hybrid TCN--Transformer model.}
    \label{fig:generalConfusion}
\end{figure}

The model achieves an accuracy of 0.9398, indicating that the majority of conjunction events are correctly classified. The recall value of 0.8125 shows that the model is able to detect most risky events, which is important for collision risk monitoring. In addition, the specificity of 0.9572 demonstrates strong performance in identifying non-risk events and limiting false alarms. The F1-score of 0.7647 reflects a balanced learning performance between detecting risky events and avoiding incorrect risk predictions. In addition, the F2-score of 0.7926 is higher than the F1-score, indicating that the model performs better in a recall-oriented evaluation, where detecting actual risky events is given more importance than minimizing false positives. Therefore, these results show that the hybrid TCN--Transformer model learns both risky and non-risk event patterns, with a stronger capability in identifying true risk events, which is particularly important for early collision risk assessment. Figure~\ref{fig:generalConfusion} presents the test set confusion matrices of the hybrid TCN--Transformer model. 

Figure~\ref{fig:ConfusionMatrix} presents the event-based confusion matrix. 112 no-risk events are correctly classified as no-risk, corresponding to 96\% of the no-risk class, while 5 no-risk events are classified as risk, corresponding to 4\%. For the risk class, 13 risky events are correctly classified as risk, corresponding to 81\%, whereas 3 risky events are classified as no-risk, corresponding to 19\%. Figure~\ref{fig:ConfusionMatrixPrefix} presents the CDM-sequence-based next-step classification result. In this evaluation, the observed CDM information is used to predict the risk state of the subsequent CDM. The model correctly classifies 834 no-risk samples as no-risk, corresponding to 94\% of the no-risk class, while 52 no-risk samples are classified as risk, corresponding to 6\%. For the risk class, 130 risky samples are correctly classified as risk, corresponding to 85\%, whereas 23 risky samples are classified as no-risk, corresponding to 15\%.

\section{Conclusion}
This study presented a CDM based learning framework for the early prediction of satellite collision probability in LEO conjunction events. The proposed methodology combines physically derived dataset construction with a hybrid TCN--Transformer model to estimate the PoC value expected in the subsequent CDM update. First, the sensitivity of CDM-related propagation parameters was analyzed using the Unscented Transform. The results showed that a $1\%$ controlled variation in the selected state and physical parameters can lead to significantly larger covariance variations, reaching approximately $30\%$--$68\%$. This confirms that the evolution of orbital uncertainty is highly sensitive to small changes in the input parameters and must be considered when constructing a PoC prediction dataset. In addition, based on the PCA findings, the raw CDM parameters were enriched with derived physical metrics representing encounter geometry, covariance dispersion, relative motion, temporal trends, and risk labels.

The generated dataset which is defined enriched conjunction data is used to train a hybrid TCN--Transformer model for next-step PoC prediction. The prediction results on the test set show that the proposed model can reproduce the general magnitude and temporal trend of future PoC values over a wide numerical range. In particular, the model successfully predicts high-risk cases on the order of $10^{-3}$--$10^{-4}$, low-risk cases on the order of $10^{-7}$--$10^{-8}$, and zero-PoC cases without introducing artificial collision risk. This indicates that the model is capable of distinguishing not only the severity level of risky conjunctions but also events with negligible or zero estimated collision probability. The event-based classification results further demonstrate the operational relevance of the framework, with an accuracy of $0.9398$, recall of $0.8125$, specificity of $0.9572$, F1-score of $0.7647$, and F2-score of $0.7926$. These results indicate that the model is capable of detecting most risky conjunction events while maintaining a low false-alarm tendency for non-risk cases. From an operational perspective, the proposed framework can be used as an additional decision-support layer in conjunction assessment workflows. When a new CDM is received, the available CDM history of the same close approach event is processed by the trained model to estimate the expected PoC of the next CDM before that update becomes available. For LEO satellites with limited maneuver capability or electric propulsion systems, such early risk information supports earlier operational planning by identifying events that may require closer monitoring, additional analysis, or preliminary maneuver preparation. Therefore, the proposed hybrid TCN--Transformer framework contributes to more timely and informed conjunction risk assessment using sequential CDM.

\section*{Acknowledgment}
This work is conducted within the context of graduate research at Istanbul Technical University and space mission research activities at TÜBİTAK UZAY. 

\section*{Future Work}
The proposed framework will be extended using an ensemble learning architecture to improve prediction robustness and generalization capability. In addition to future PoC estimation, further studies will investigate the prediction of other CDM-related parameters, including relative motion, covariance based uncertainty indicators, and encounter geometry metrics, to provide a more comprehensive early conjunction risk assessment framework.

\bibliographystyle{AAS_publication}   
\bibliography{references}   

\end{document}